\documentclass[runningheads]{llncs}

\usepackage{eccv}

\usepackage{eccvabbrv}

\usepackage{graphicx}
\usepackage{booktabs}

\usepackage{cite}
\usepackage[table]{xcolor}
\usepackage{colortbl}
\usepackage{booktabs}
\usepackage{amssymb}
\usepackage{textcomp}
\usepackage{marvosym}
\usepackage{wrapfig}
\usepackage{float}

\usepackage[accsupp]{axessibility}  

\usepackage{hyperref}

\usepackage{orcidlink}

\begin{document}

\title{ASSCG: Just-Right Gating over Chattering for Fast–Slow LLM Planning in Autonomous Driving} 

\titlerunning{ASSCG: Just-Right Gating for LLM Planning}

\newcommand{\corresponding}{\textsuperscript{*}}
\author{Sining Ang\inst{1,2} \and
Yuan Chen\inst{3} \and
Liu Haiyan\inst{4} \and
Xuanyao Mao\inst{4} \and
Jason Bao\inst{4} \and
Xuliang\inst{4} \and
Bingchuan Sun\inst{4}\corresponding \and
Yan Wang\inst{1}\corresponding}

\authorrunning{S.~Ang et al.}

\institute{Institute for AI Industry Research (AIR), Tsinghua University\\
\email{angsn@mail.ustc.edu.cn, wangyan@air.tsinghua.edu.cn} \and
Department of Automation, University of Science and Technology of China \and
Beijing University of Aeronautics and Astronautics\\
\email{chenyuan1@buaa.edu.cn} \and
Lenovo Group Limited\\
\email{\{liuhy46,maoxy6,jbao,xuliang18,sunbc1\}@lenovo.com}}

\maketitle
\begingroup
\renewcommand{\thefootnote}{\fnsymbol{footnote}}
\footnotetext[1]{Corresponding authors.}
\endgroup

\begin{abstract}
  Large language models (LLMs) can improve autonomous driving planning but are costly to query online, and existing fast–slow planners often rely on hand-designed triggering rules that either over-call the slow system or call it at the wrong times. We formulate slow-system invocation as a resource-aware sequential decision problem and propose the Adaptive Slow-System Control Gate (ASSCG), which makes frame-level Query/Cache/Drop decisions to refresh, reuse, or suppress slow guidance. ASSCG uses an RWKV backbone for efficient long-horizon gating and is trained with supervised fine-tuning followed by GRPO-style compute-aware reinforcement fine-tuning. We apply ASSCG to two different fast–slow architectures: (i) AsyncDriver on nuPlan Hard20 closed-loop evaluation, where ASSCG improves score to 67.28 (+2.28) while reducing average end-to-end inference latency by $\sim$60\%; and (ii) a RecogDrive-based dual system that we build by replacing its original VLM-2B module with a lightweight ViT-based fast planner and adding an LLM slow planner, evaluated on NAVSIM, where ASSCG achieves 91.4 PDMS (+0.6) and increases average speed by $\sim$25\%. The project page, including video visualizations and additional results, is available at \url{https://williamxuanyu.github.io/asscg/}.

\end{abstract}

\section{Introduction}
\label{sec:intro}

Motion planning is a core component of autonomous driving. Despite notable progress in learning-based planning and standardized benchmarks~\cite{zhang2020multi,nuplan}, modern planners still struggle in complex, dynamic, long-tail interactions~\cite{hu2024solving,cheng2024pluto,hallgarten2023prediction,cheng2024rethinking}. In parallel, large language models (LLMs) exhibit strong commonsense reasoning and cross-domain transfer; with instruction tuning, they can internalize traffic rules and scenario priors to provide more context-aware guidance~\cite{fu2024drive,han2025dme,cui2024receive,chen2024driving,chen2023towards}. This has motivated LLM-augmented planning.

However, LLM inference latency and compute cost make per-step deployment difficult. A practical compromise is the decoupled fast--slow paradigm: a real-time fast planner runs every frame while an LLM-based slow module provides high-level guidance at a lower frequency~\cite{chen2024asynchronous,tian2024drivevlm,tang2025vlmplanner}. Yet in practice, peak performance often requires querying the slow module nearly every frame; reducing the query rate typically degrades results. More importantly, adding a slow LLM module does not always help and can even hurt in certain segments (Fig.~\ref{fig:fig2}), raising a central question: \emph{when should the system rely on the fast planner, and when should it consult (or ignore) the slow module, to maximize closed-loop performance under a compute budget?}

\begin{wrapfigure}[27]{r}{0.4\linewidth}
    \centering
    \includegraphics[width=\linewidth, height=\textheight, keepaspectratio]{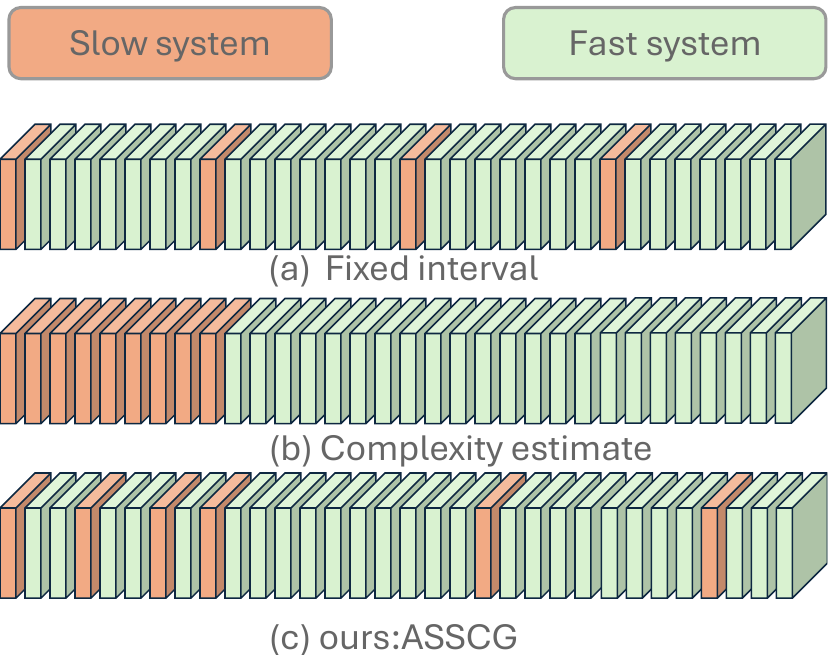}
    \caption{Common fast--slow coordination strategies: (a) fixed-interval triggering ignores temporal variation, wasting compute in easy periods and missing critical moments; (b) difficulty/complexity-based triggering relies on imperfect proxies that not align with the value of slow reasoning, leading to unnecessary oscillation and mis-timed queries; (c) ours (ASSCG), a learned gate that makes frame-level \textsc{Query}/\textsc{Cache}/\textsc{Drop} decisions to balance planning quality and resource cost.}
    \label{fig:1}
\end{wrapfigure}

Two coordination strategies are widely used (Fig.~\ref{fig:1}): (i) fixed-interval triggering and (ii) heuristic difficulty-based triggering. Fixed schedules ignore intra-scene variability, leading to over-calling in easy segments and under-calling in hard ones. Difficulty-based gating requires defining and estimating ``scene complexity'', yet such proxies do not necessarily align with the \emph{marginal utility} of invoking the slow module at a particular time. Empirically, VLMPlanner evaluates multiple gating variants (rule-based and learned complexity criteria) across different settings and hyperparameters, but none improves over the always-querying baseline in terms of final score; the best accuracy is still achieved by per-frame slow-module invocation, while the best efficiency–accuracy trade-off comes from a learned complexity gate that slightly reduces score but significantly improves throughput~\cite{tang2025vlmplanner}. Moreover, most existing schemes operate at the scene level with fixed fast/slow frequencies per class, overlooking fine-grained temporal dynamics within a single episode.

We propose the \textbf{A}daptive \textbf{S}low-\textbf{S}ystem \textbf{C}ontrol \textbf{G}ate (ASSCG), an architecture-agnostic control interface that adaptively schedules and regulates a slow module at the frame level. Here, architecture-agnostic means that ASSCG only assumes a fast branch, an optional slow branch, and a cached guidance representation; the gate is still trained or lightly adapted for the target planner and benchmark. We formulate slow-module control as a sequential decision problem with query costs under partial observability (POMDP/SMDP), and recast it as sequence generation: the controller predicts a gating token per frame. This LLM-style formulation enables standard SFT-to-RL fine-tuning. Concretely, we implement ASSCG with an RWKV backbone~\cite{peng2023rwkv} and fine-tune it with a compute-aware GRPO-style algorithm~\cite{shao2024deepseekmath}. On nuPlan Hard20 closed-loop evaluation, integrating ASSCG into AsyncDriver (hereafter referred to as \emph{AdaptiveAsyncDriver}) reduces average end-to-end latency by about 60\% while improving the score by +2.28 (to 67.28). We further demonstrate the applicability of the same gating principle on a separate RecogDrive-based fast--slow instantiation evaluated on NAVSIM, where it improves PDMS by +0.6 and increases average speed by $\sim$25\%. Our contributions are:
\begin{enumerate}
    \item We analyze fast--slow collaboration and introduce \emph{effective} and \emph{failure} intervals to characterize when slow guidance helps or hurts over time.
    \item We present ASSCG, a frame-level gate trained with GRPO-style reinforcement fine-tuning to optimize closed-loop metrics under query costs.
    \item We evaluate ASSCG on two representative fast--slow instantiations and benchmarks: AsyncDriver on nuPlan closed-loop Hard20 and a RecogDrive-based fast--slow planner on NAVSIM, demonstrating improved performance--efficiency trade-offs across architectures and evaluation protocols.
\end{enumerate}

\section{Related Work}
\subsection{Language-augmented motion planning with LLMs}
Large language models (LLMs) exhibit strong generalization and reasoning after pre-training on massive corpora, motivating their use in autonomous driving decision-making. Recent works encode ego state, surrounding agents, and map context as linguistic descriptions or BEV-style representations for LLM/VLM-based scene understanding and trajectory guidance~\cite{zheng2024planagent,sharan2023llm,yao2024calmm}. For example, PlanAgent~\cite{zheng2024planagent} structures scenes in BEV to guide a base planner, and LLM-ASSIST~\cite{sharan2023llm} integrates LLM reasoning with rule-based modules.

Purely language-mediated interfaces, however, face an expressivity--latency mismatch for time-critical control: limited context length and the difficulty of encoding precise continuous states can reduce reliability. Multimodal approaches---e.g., DrivingWithLLM~\cite{chen2024driving}, DriveGPT4~\cite{xu2024drivegpt4}, and RAGDriver~\cite{yuan2024rag}---align vectorized or image/video inputs with language to enrich interpretation, but translating language outputs into stable closed-loop control remains challenging. Another line couples LLMs more tightly with low-level controllers for closed-loop driving, such as LMDrive~\cite{shao2024lmdrive} and LanguageMPC~\cite{wu2025language}, yet these designs still rely on frequent LLM inference or serial decoding, stressing real-time responsiveness.

\subsection{Fast--slow (dual-system) planners and adaptive LLM invocation}
To better satisfy real-time constraints while leveraging LLM reasoning, fast--slow planners decouple a real-time fast planner from an LLM-based slow module that provides high-level guidance at a lower frequency~\cite{chen2024asynchronous,tian2024drivevlm}. AsyncDriver~\cite{chen2024asynchronous} caches slow guidance and reuses it across frames, amortizing cost but typically requiring careful scheduling to avoid stale or mis-timed interventions under non-stationary traffic.

A number of works study \emph{when} to invoke the slow module. VLMPlanner~\cite{tang2025vlmplanner} introduces CAI-Gate to adjust query frequency using simple/complex judgments from a VLM; its reported gains are marginal under multiple gating criteria, suggesting that generic scene-difficulty proxies may not reliably track the \emph{value} of invoking the slow planner. Uncertainty-based triggering (e.g., FASIONAD~\cite{qian2024fasionad}) calls the slow module under predicted waypoint uncertainty, but uncertainty calibration in interactive long-tail scenes can be brittle. Task-specific systems (e.g., Chameleon~\cite{zhang2025chameleon}) allocate extra slow reasoning for detected edge cases, often requiring curated traces or additional annotations.

\textbf{Recent concurrent work.} AdaDrive~\cite{zhang2025adadrive} proposes an adaptive slow--fast framework that learns \emph{when} to activate LLM assistance via an adaptive activation loss, and further introduces a continuous fusion mechanism that scales the LLM contribution based on scene complexity and prediction confidence. Our work is complementary but differs in several aspects: (i) we learn an \emph{architecture-agnostic}, \emph{frame-level} discrete gate with explicit \textsc{Query}, \textsc{Cache}, and \textsc{Drop} actions, where \textsc{Drop} enables intentionally suppressing slow guidance when it is likely to be harmful; (ii) we train the gate with supervised pretraining followed by compute-aware RL fine-tuning to directly optimize task-level driving metrics under an explicit query-cost constraint; and (iii) rather than scaling LLM influence using confidence as a primary signal, we focus on \emph{when to consult} and \emph{whether to trust} the slow module---in our setting, the fast planner already performs probability-based trajectory aggregation (GameFormer~\cite{huang2023gameformer}), which can make confidence among shortlisted candidates less discriminative for further modulation. Overall, our goal is to learn just-right \emph{timing} and \emph{usage} of slow reasoning under resource constraints.

\begin{figure*}[ht]
    \centering
    \includegraphics[width=1\linewidth]{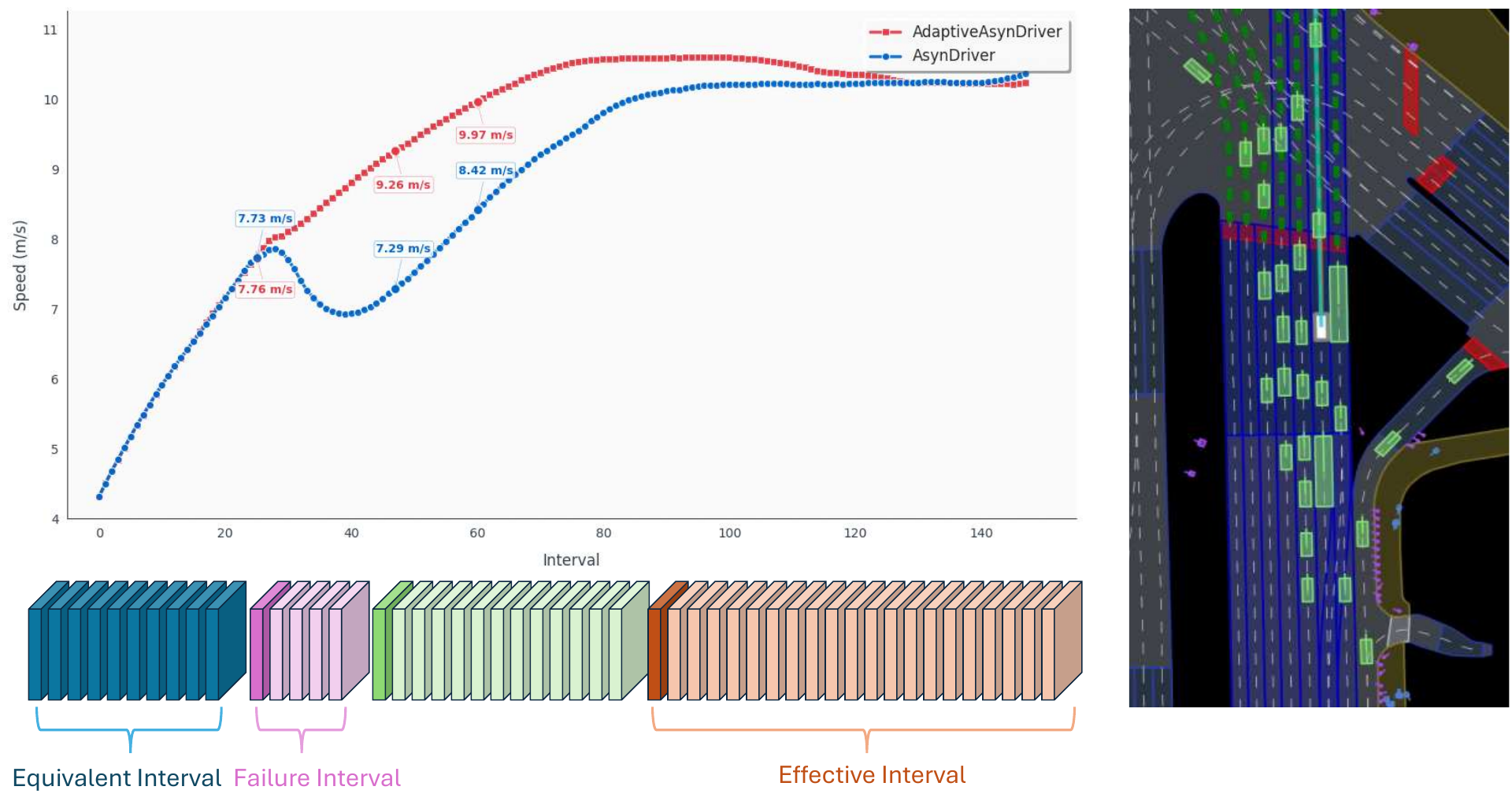}
    \includegraphics[width=1\linewidth]{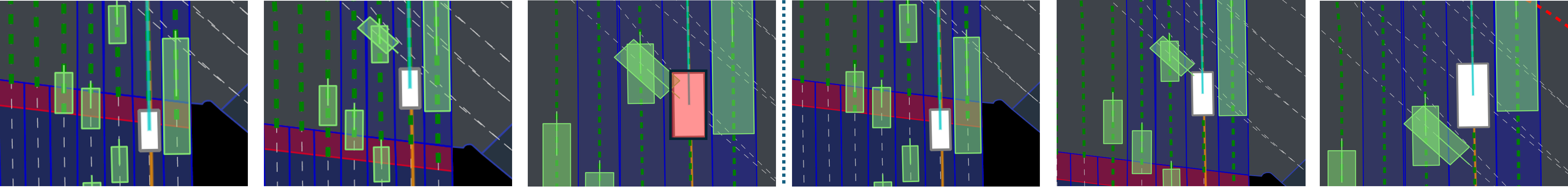}
    \caption{Straight-driving case study comparing AsyncDriver (always querying the slow planner) and AdaptiveAsyncDriver with ASSCG (querying only at frames 0, 22, and 89). Despite far fewer slow-system calls, ASSCG avoids a collision and achieves better closed-loop behavior. Frames 0--25 form an Equivalent Interval (EI); frames 25--40 reveal a Failure Interval (FI) where slow guidance is harmful; and frames 1--25 and 90--149 are Effective Intervals (EfI) for guidance issued at frames 0 and 89, respectively. Additional qualitative visualizations are provided in the supplementary material.}
    \label{fig:fig2}
\end{figure*}


\section{Analysis}
\label{sec:analysis}
\subsection{Setup and assumptions}
Following the AsyncDriver paradigm, we treat the LLM-based slow system as a high-level intent generator that does not need to be queried at every frame. The real-time fast planner consumes and caches the most recent slow-system guidance and continues planning until the next refresh. Concretely, let $q_k$ denote the frames at which the slow system is invoked, $z_{q_k}$ the resulting high-level guidance (e.g., latent features, waypoints, or policy logits), and $u_t$ the low-level control produced by the fast planner at frame $t$ using the latest cached guidance $z_{q_k}$, where $q_k \leq t < q_{k+1}$.

\subsection{Concepts and operational definitions} 
To analyze when and how the slow system should be queried, we introduce three intervals and provide operational criteria for measurement in practice.

\textbf{Equivalent Interval (EI).} A post-query time window in which re-querying the slow system would be redundant because the incremental information is negligible. Operationally, for $t \in [q_k, q_{k+1})$:
(i) $similarity(z_{q_k}, z_{t'}) \geq \tau_{sim}$ for all $t'$ in the window, where similarity can be cosine similarity in the slow-system latent space or an embedding distance after a projection; or (ii) the induced control difference is negligible, e.g., $||u_t(z_{q_k}) - u_t(z_{q_{k'}})|| \leq \tau_{ctrl}$ for a hypothetical re-query at frame  $q_{k'} = t$.
Intuitively, EI indicates that the slow guidance remains valid and “fresh,” and repeated slow queries would not change the planned motion in a meaningful way.

\textbf{Effective Interval (EfI).} The span during which the current slow guidance is actually used by the fast planner before being refreshed, i.e., $[q_k, q_{k+1}]$. By definition EfI is the cache-validity window; in practice it may be longer or shorter than EI. When $EfI \approx EI$, caching is efficient; when $EfI \gg EI$, stale guidance can accumulate; when $EfI \ll EI$, compute is wasted.

\textbf{Failure Interval (FI).} A time window in which invoking the slow system degrades downstream performance relative to not invoking it. Operationally, for a candidate call at $t$:
$\Delta J_t = J(u_t \mid \text{call at } t) - J(u_t \mid \text{no call at } t) < -\tau_{\text{perf}}$ ,
where $J$ aggregates safety and efficiency objectives (e.g., collision, lane-keeping, progress, comfort). Typical causes include misestimated distances, occlusion-induced hallucinations, or misinterpretation of local interaction rules. FI emphasizes that more frequent slow reasoning is not always better.

These definitions align with a value-of-information perspective: query the slow system only when the expected improvement in $J$ exceeds the query cost and the risk of FI.

\subsection{Case study and empirical observations}
Fig.~\ref{fig:fig2} illustrates a straight-driving episode comparing AsyncDriver (querying the slow module every frame) and our approach (querying only at frames 0, 22, and 89). Despite making only three slow-module calls, our gate avoids a collision and outperforms the always-query baseline. Notably, frames 0--25 behave as an Equivalent Interval (re-querying yields negligible change), while frames 25--40 expose a Failure Interval where slow guidance is detrimental in this instance. This example motivates the need for a gate that can both \emph{reuse} and \emph{suppress} slow guidance depending on context.

\paragraph{Beyond a single case: fixed-schedule grid search across scenario types.}
To check whether the above behavior is isolated, we run a grid search of \emph{fixed} slow-module query schedules on nuPlan Hard20, separately for each of the 14 scenario types. We evaluate fixed query stride $K \in \{1,3,5,10,20,50,100\}$ (query every $K$ frames) and a \textit{Never-Query} setting that disables the slow module. Across types, the best fixed schedule varies substantially---from always querying ($K{=}1$) to rarely querying ($K{=}50$), and in one case \textit{Never-Query} performs best. Notably, this grid search assigns a \emph{single} stride to all episodes within a type; the optimal stride can still differ across episodes within the same type due to intra-scene temporal dynamics. This further motivates learning a frame-level policy that adapts query decisions online. Detailed per-type results are provided in Appendix~\ref{app:gridsearch_stride}.

\subsection{Implications for gating policy design}
The analysis motivates a frame-level gating policy with three objectives:
\begin{enumerate}
\item Minimize redundant slow queries within Equivalent Intervals by reusing previously issued slow guidance and deferring new slow calls while the scene and control remain within the same equivalence window.
\item Suppress slow queries in segments unsuitable for LLM involvement (e.g., Failure Intervals), thereby avoiding adverse interventions.
\item Extend Effective Intervals by continuing to use cached slow guidance until a refresh is needed, subject to a maximum reuse horizon.
\end{enumerate}

ASSCG operationalizes these objectives by casting the slow-trigger decision as a sequential problem under a constrained query budget. It seeks to maximize $J$ by invoking the slow system only when the expected marginal benefit justifies the query cost; otherwise, it reuses cached guidance and leverages the fast planner. Implementation details are provided in the following section.

\section{Methodology}

\begin{figure*}[ht]
    \centering
    \captionsetup{skip=2pt}
    \includegraphics[width=1\linewidth]{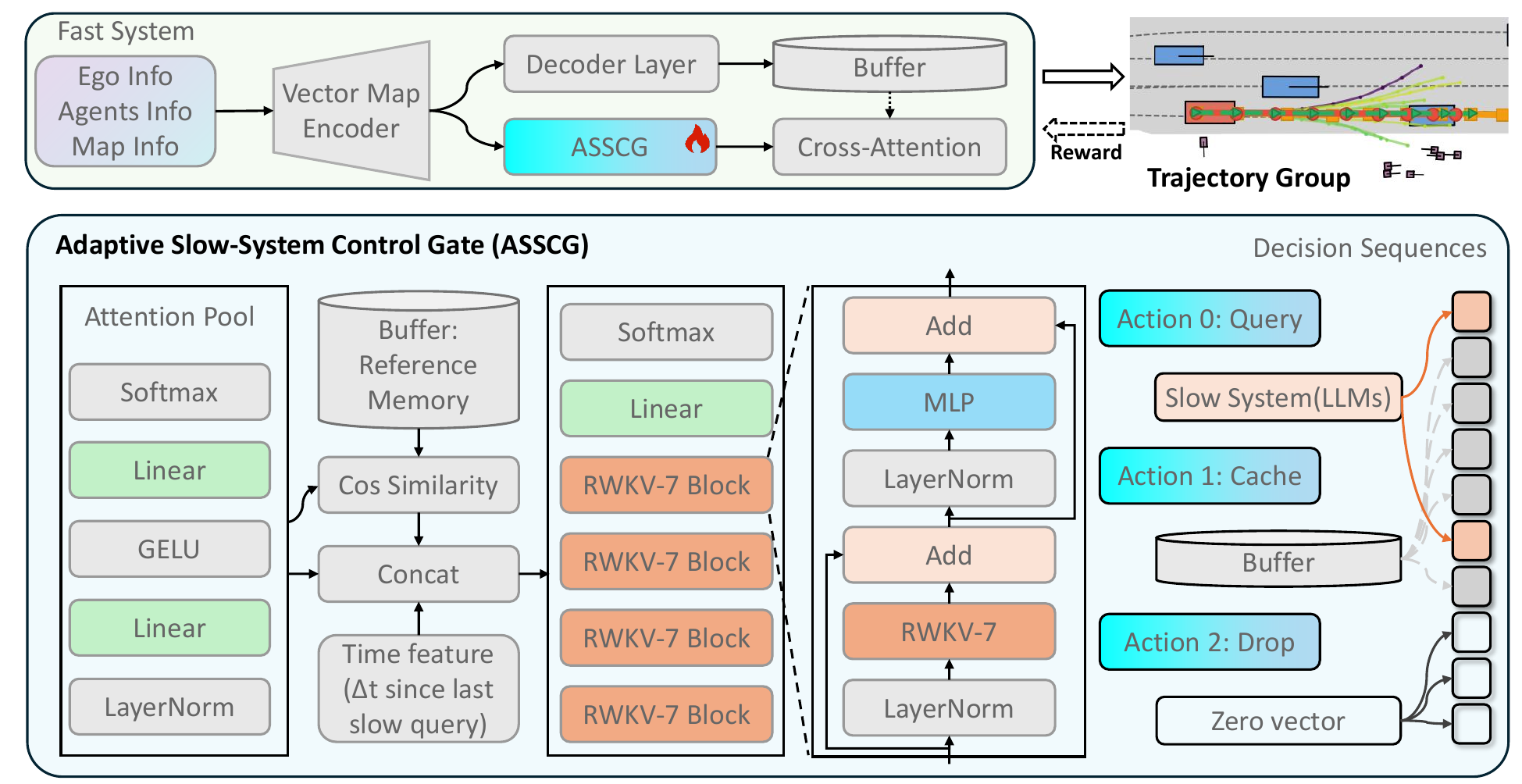}
    \caption{Overview of our framework. We couple a GameFormer-based~\cite{huang2023gameformer} fast planning system with an LLM-based slow system, coordinated by an Adaptive Slow-System Control Gate (ASSCG). At each frame, encoded vector-map (and ego/agent features) is fed to both the fast decoder and ASSCG. ASSCG outputs a discrete gating action: \textbf{Query} invokes the slow system to refresh a reference-memory buffer, \textbf{Cache} reuses the buffer without querying the slow system, and \textbf{Drop} ignores the buffer (zero vector) to prevent stale memory from affecting planning. Conditioned on the gating decisions, the fast decoder fuses fast features with the (optional) buffered reference via cross-attention and outputs a trajectory for one rollout. A \emph{trajectory group} is obtained by sampling multiple rollouts, whose scores are converted to rewards to optimize the per-frame gating policy with group-based RL (GRPO). Inside ASSCG, a lightweight attention pool aggregates cues, which are compared with the buffer via cosine similarity and concatenated with time feature before an RWKV backbone predicts the gating policy.}
    \label{fig:main}
\end{figure*}

\subsection{Overview}
Our framework follows a decoupled fast--slow architecture. A learning-based real-time planner (fast system) runs every frame, while an LLM-based planner (slow system) is invoked on demand to provide high-level guidance that is cached in a buffer and fused into the fast planner via cross-attention. The Adaptive Slow-System Control Gate (ASSCG) decides at each frame whether to (i) \textsc{Query} the slow system and refresh the buffer, (ii) \textsc{Cache} and reuse the buffered guidance, or (iii) \textsc{Drop} the guidance by substituting a null feature.

\paragraph{Training pipeline and parameter freezing.}
Following AsyncDriver, the underlying fast--slow planner is trained in three stages: (i) we first train the fast planner (GameFormer-style) independently; (ii) we then fine-tune the slow LLM (LLaMA2-13B) with LoRA for driving-domain adaptation; and (iii) we subsequently train the full fast--slow planner end-to-end while freezing the LLM weights, so that the fast planner and fusion modules adapt to the LLM guidance without updating the LLM itself. After inserting ASSCG, we perform an additional, separate training stage for the gate: we freeze \emph{all} parameters of the original AsyncDriver and optimize only ASSCG. This isolates the effect of improved slow-system scheduling and selective usage from further planner adaptation.

\subsection{Model architecture}
The core dual-system architecture follows the AsyncDriver paradigm. Let $m_t$ denote the vectorized map-and-agent features produced by a Vector Map Encoder at frame $t$; the fast planner’s decoder consumes $m_t$ to predict low-level control and a short-horizon trajectory. When available, cached slow guidance $b_t$ from the buffer is injected into the decoder through a cross-attention path. The slow system (LLM) is queried only when gated “on,” producing high-level intent $z_t$ (e.g., latent features, waypoints, or policy logits), which is written to the buffer.

Concretely, in a decoder block $l$, we extend the standard attention with an instruction-injection:
$${s}^{l + 1} = g \cdot  \operatorname{softmax}\left( \frac{Q{K}_{\widehat{h}}^{T}}{\sqrt{C}}\right) {V}_{\widehat{h}} + \operatorname{softmax}\left( \frac{Q{K}_{{s}^{l}}^{T}}{\sqrt{C}}\right) {V}_{{s}^{l}}$$
where $h$ is the instruction feature projected from the slow guidance (last-token hidden state after a small adapter), ${s}^{l}$ is the scene feature at block $l$, and $g$ is a learnable gate.

Per iteration, we obtain vectorized ego\&agent histories and map elements in the ego-centric frame, similar to GameFormer-style encoders. The Vector Map Encoder produces tokenized features for lanes, topology, and dynamic agents. These tokens drive the fast planner, while ASSCG decides whether to reuse features in the buffer, refresh the buffer with new LLM-derived features, or bypass LLM features altogether.

\subsection{Buffering and guidance representation}
\textbf{Buffer content.} The buffer stores a compact slow-guidance representation $b = Proj(z)$, along with map tokens and the current timestamp (frame).
\textbf{Read/write policy.} When the slow system is queried at frame $t$, we overwrite $b \leftarrow Proj(z_t)$. When not queried, we reuse the previous $b$. When the gate predicts “drop,” we expose a null vector $b_0$ to the planner, effectively disabling slow guidance.

\subsection{ASSCG: Adaptive Slow-System Control Gate}
The gating policy must determine how the planner should interact with the buffer at every frame— whether to refresh it with new slow guidance, keep using the existing contents, or intentionally ignore it to avoid harmful interventions. This decision depends on the temporal context, cached features, and the learned scheduling strategy. 

\textbf{Problem framing:} At each frame t, ASSCG produces an action $a_t \in \{0, 1, 2\}$:
\begin{itemize}
    \item 0 = Query (invoke the slow system and refresh the buffer),
    \item 1 = Cache (reuse the buffer as-is),
    \item 2 = Drop (ignore the buffer by substituting a null feature).
\end{itemize}
This realizes our interval analysis: \textsc{Cache} extends effective intervals, \textsc{Query} marks the start of a new interval, and \textsc{Drop} suppresses suspected failure intervals. The first frame of every episode is forced to \textsc{Query} to initialize the buffer, so \textsc{Cache} is never applied without valid slow guidance.

\textbf{Inputs to the gate:} Current scene embedding $x_t$ from the Vector Map Encoder (token sequence); Reference memory $b_{ref}$ from the buffer (the last slow-guidance embedding); Cosine similarity $s = cos(Pool(x_t), b_{ref})$, where $Pool$(·) is a lightweight attention pooling that summarizes $x_t$; Time feature is $\Delta t$ since the last slow query (normalized via logarithmic scaling). The gate concatenates $[Pool(x_t),s, \Delta t]$ as its input feature.

\textbf{Attention pooling:} We employ a compact attention-pooling block (Linear–GELU–Linear–Softmax with LayerNorm) to summarize the token sequence. It mirrors the Map Adapter design in AsyncDriver for map embeddings, ensuring architectural consistency and integrating vectorized map features with minimal computational overhead.

\textbf{Why RWKV for gating.}
ASSCG must make low-latency, frame-level decisions while tracking long-horizon temporal context (up to an entire episode). We adopt RWKV~\cite{peng2023rwkv} for two reasons. First, its recurrent time-mixing can be viewed as maintaining a compact, recursively updatable latent state, which matches the intuition that the gate benefits from an internal belief state under partial observability. Second, RWKV provides a favorable deployment profile for online decision-making with long context: after state caching, it has per-step linear time and constant memory, avoiding the quadratic attention cost of Transformers as the episode length grows. We empirically compare RWKV with a parameter-matched Transformer alternative in the ablation study (\S\ref{sec:ablation}).

\textbf{RWKV backbone and head:}
The pooled feature and auxiliaries are fed to a 4-layer RWKV-7 stack. A linear classification head with softmax outputs $\pi(a_t \mid s_t)$ over \{\textsc{Query}, \textsc{Cache}, \textsc{Drop}\}. As discussed above, RWKV offers an efficient long-horizon gating backbone; quantitative accuracy/latency comparisons against Transformer-based gating are reported in the ablations.

\section{Data Collection and Training Protocol}

\subsection{Datasets}
\label{sec:datasets}
\paragraph{Scenario types and splits.}
We follow the nuPlan Challenge 2023 closed-loop setting and evaluate on the 14 official scenario types~\cite{nuplan}. For a fair comparison with prior fast--slow planners, we use the \textit{Hard20} test split released by AsyncDriver~\cite{chen2024asynchronous} and VLM-planner~\cite{tang2025vlmplanner} (same scene IDs and evaluation protocol), and report results on this fixed split throughout.

\paragraph{SFT training pool.}
From the training split, we randomly sample 60 scenes per scenario type (14 types). We collect training rollouts by executing a set of fixed slow-module query schedules in the nuPlan simulator, using query stride $K \in \{1,3,5,10,20,50,100\}$ (query every $K$ frames) and a \textit{Never-Query} setting that disables the slow module. For each rollout, we log per-frame encoder features, the slow-module output (when queried), the gate action token, and metric components of the nuPlan score (Eq.~\ref{equ:nuplan score} in Appendix~\ref{sec:sup_metric}).

To obtain supervision for the gate, we select, \emph{for each scene}, the rollout with the highest closed-loop score among the above schedules, and use its per-frame \textsc{Query}/\textsc{Cache}/\textsc{Drop} action sequence as pseudo labels for supervised fine-tuning (SFT).

\paragraph{RL data preparation.}
We construct an RL dataset by rolling out the SFT-initialized gate policy in the nuPlan simulator and then training from the logged trajectories. The gate action space has $A=3$ actions (\textsc{Query}, \textsc{Cache}, \textsc{Drop}); each episode spans the full closed-loop horizon.

We adopt an offline RL workflow: using the frozen SFT policy \(\pi_0\), each training scene is rolled out \(R=10\) times with different random seeds. Actions are selected via entropy-shaped temperature scaling with \(\varepsilon\)-mixing. For a state \(s\), let \(p=\mathrm{softmax}(\mathrm{logits}(s))\) over the action set \(A\), and define  
\begin{equation}
\begin{aligned}
&H(s)=-\sum_{i=1}^{A} p_i \log p_i,\quad H_{\mathrm{norm}}(s)=\frac{H(s)}{\log A}\in[0,1],\quad \\
&T(s)=T_{\mathrm{low}}+\bigl(T_{\mathrm{high}}-T_{\mathrm{low}}\bigr)\bigl(H_{\mathrm{norm}}(s)\bigr)^{\alpha},\quad \\
&\varepsilon(s)=\varepsilon_{\mathrm{low}}+\bigl(\varepsilon_{\mathrm{high}}-\varepsilon_{\mathrm{low}}\bigr)\bigl(H_{\mathrm{norm}}(s)\bigr)^{\alpha}.
\end{aligned}
\label{eqn-H}
\end{equation}

With $p_T(a\mid s)=\mathrm{softmax}(\mathrm{logits}(s)/T(s))$ and $U(a)=1/A$, the behavior distribution is

\begin{equation}
\begin{aligned}
&p_{\mathrm{mix}}(a\mid s)=(1-\varepsilon(s))\,p_T(a\mid s)+\varepsilon(s)\,U(a),\quad \\
&a\sim \mathrm{Cat}\bigl(p_{\mathrm{mix}}(\cdot\mid s)\bigr)
\end{aligned}
\label{eqn-p}
\end{equation}
using $T_{\mathrm{low}}{=}0.9$, $T_{\mathrm{high}}{=}1.6$, $\varepsilon_{\mathrm{low}}{=}0.02$, $\varepsilon_{\mathrm{high}}{=}0.12$, $\alpha{=}1.0$. We force a \textsc{Query} at the first step of each episode to initialize the buffer. The resulting static dataset $D$ logs per-step $s$ (encoder features), $a$, reward metrics for offline reward reconstruction (e.g., TTC, Lim, Comf, Prog; see in Appendix~\ref{sec:sup_metric}), the behavior log-probability $\log b=\log p_{\mathrm{mix}}(a\mid s)$, as well as $H_{\mathrm{norm}}(s)$, $T(s)$, $\varepsilon(s)$, and metadata (done, episode\_id, step\_idx, timestamps).

\subsection{Training protocol}
We train ASSCG in two stages: supervised fine-tuning (SFT) on pseudo labels, followed by compute-aware RL fine-tuning.

\paragraph{Stage I: Supervised fine-tuning.}
Over the action set $a_t \in \{\textsc{Query},\textsc{Cache},\textsc{Drop}\}$, we train a 3-way classifier using the per-frame action sequence from the highest-scoring fixed-schedule rollout of each scene (\S\ref{sec:datasets}). We minimize cross-entropy with class-balancing to avoid the degenerate \textsc{Cache}-only policy. We use AdamW (lr $3{\times}10^{-4}$, weight decay 0.01) with cosine decay and 10\% warmup, label smoothing 0.05, and gradient-norm clipping at 1.0.

\paragraph{Stage II: Compute-aware RL fine-tuning.}
The nuPlan closed-loop score (Eq.~\ref{equ:nuplan score} in Appendix~\ref{sec:sup_metric}) is computed as a weighted sum of per-episode metrics, normalized by the sum of weights $W \!=\! 16$, and multiplied by gating terms (e.g., collision/drivable/direct). Since Eq.~\ref{equ:nuplan score} in Appendix~\ref{sec:sup_metric} is episode-level, it provides sparse supervision. We introduce dense intermediate feedback via potential-based shaping while preserving alignment with the official weighting. Let $W_{\mathrm{np}} \!=\!11$ denote the total weight of the non-progress terms (\textit{TTC}, \textit{Lim}, \textit{Comf}). Define
\begin{equation}
\begin{aligned}
N_t = 5\,\mathrm{TTC}_t + 4\,\mathrm{Lim}_t + 2\,\mathrm{Comf}_t,\quad
G_t = \mathrm{Coll}_t \times \mathrm{Drivable}_t\times\mathrm{Direct}_t,
\end{aligned}
\end{equation}
and the normalized potential
\begin{equation}
\Phi_t = \frac{1}{W}\, N_t \, G_t \;=\; \frac{W_{\mathrm{np}}}{W}\cdot
\underbrace{\Big(\frac{N_t}{W_{\mathrm{np}}}G_t\Big)}_{\psi_t \in [0,1]}.
\end{equation}
The per-step reward is
\begin{equation}
r_t =
(\gamma \Phi_{t+1} - \Phi_t)
\;-\;
\lambda_{\mathrm{call}}\cdot \mathbb{I}[a_t=\textsc{Query}],
\label{eq:reward_main}
\end{equation}
with an additional terminal progress settlement
\begin{equation}
r_T \leftarrow r_T + \frac{W-W_{np}}{W}\,\mathrm{Prog}_T\,G_T .
\end{equation}
With $\gamma \approx 1$, the shaping term telescopes, yielding a constant-shifted version of the non-progress part of Eq.~\ref{equ:nuplan score} in Appendix~\ref{sec:sup_metric}, and the terminal term restores the progress component. This yields an objective consistent with the official evaluation but with denser learning signal. The compute-awareness is explicit in the objective: every \textsc{Query} action receives a penalty controlled by $\lambda_{\mathrm{call}}$, while \textsc{Cache} and \textsc{Drop} avoid this cost. Increasing $\lambda_{\mathrm{call}}$ shifts the policy toward longer effective query intervals, trading slow-module usage against closed-loop score in a controlled way.

We fine-tune the policy using a GRPO-style clipped policy-gradient objective with (i) off-policy correction using logged behavior probabilities, (ii) KL regularization to the SFT reference policy $\pi_{\mathrm{ref}}$, and (iii) an entropy bonus for exploration. Concretely, with importance ratio $\rho_t = \pi_\theta(a_t|s_t)/b(a_t|s_t)$ truncated by $\rho_{\max}$, we apply PPO-style clipping with range $\epsilon$; full objective and advantage estimation details are provided in the supplementary material.

\paragraph{Hyperparameters.}
Unless otherwise stated: $\gamma{=}0.99$, GAE $\lambda{=}0.95$, $\lambda_{\mathrm{call}}{=}0.001$, clip $\epsilon{=}0.2$, $\rho_{\max}{=}10$, entropy coefficient $\eta{=}0.01$, value loss coefficient $\lambda_V{=}0.5$, and KL coefficient $\beta$ is set to $0.01$. We use AdamW with lr $1{\times}10^{-4}$ and gradient-norm clipping at 1.0.

\section{Experiments}
\subsection{Experimental Setup}

\paragraph{General settings.}
Unless otherwise stated, all reported scores and inference latencies are measured with single-GPU inference on an NVIDIA A30. To reduce variance from RL fine-tuning, we report the average over three runs with different random seeds.

\paragraph{nuPlan setup.}
We evaluate in the nuPlan closed-loop reactive setting: non-ego agents follow IDM-based planners and respond to the ego’s actions. The simulation runs at 10\,Hz, and each planning iteration predicts an 8\,s trajectory horizon. We use the official nuPlan closed-loop metrics; definitions and the composite score formula are summarized in Appendix~\ref{sec:sup_metric}. For fairness with prior work, we use the fixed Hard20 split from AsyncDriver \& VLMplanner (same scene IDs).

\paragraph{NAVSIM setup.}
We also evaluate on NAVSIM, a planning-oriented benchmark with a non-reactive (pseudo closed-loop) simulator. Each scene requires one-shot prediction of a 4\,s future trajectory (8 waypoints at 0.5\,s). We report the official PDMS metric (Appendix~\ref{app:navsim}).

\subsection{Results on nuPlan (AsyncDriver-based Fast--Slow Planner)}
On nuPlan Hard20, integrating ASSCG into AsyncDriver improves the overall score to 67.28 (+2.28 absolute) while reducing compute by lowering the slow-module query rate. Detailed per-scenario-type results are provided in Appendix~\ref{app:pertype_results}. Overall, we observe improvements on most scenario types, with the largest gains on lane-change and intersection-related categories, and only minor regressions on two types (type0 and type7), which are within a small margin.

In addition to the overall score, Table~\ref{tab:mainres} reports a breakdown of nuPlan metric components. ASSCG improves drivable-area and driving-direction compliance as well as TTC, indicating that the gain is primarily driven by safer and more stable closed-loop behavior.

Beyond accuracy, ASSCG yields consistent efficiency benefits. Table~\ref{table2} compares three settings enabled by the asynchronous fast--slow design of AsyncDriver: always querying the slow module, a fixed-stride schedule that queries once every 5 frames, and ASSCG. The 5-frame fixed schedule has nearly the same average latency as ASSCG (0.32 s/frame), providing a latency-matched baseline. Under this comparable efficiency budget, ASSCG improves the score from 64.27 to 67.28 (+3.01), indicating that the gain comes primarily from \emph{better timing and selective suppression} of slow-module calls rather than merely reducing their frequency. On the Hard20 test split, ASSCG achieves an average effective query interval of 5.26 frames, corresponding to a slow-query rate of roughly 19.0\% of frames. This is close to the fixed 5-frame schedule in latency, but differs in timing and in whether cached slow guidance should be trusted or dropped.

\begin{table*}[tb]
  \caption{nuPlan Hard20 closed-loop evaluation: score breakdown.
  We report the overall score and the average values of metric components used by the official nuPlan score (Eq.~\ref{equ:nuplan score} in Appendix~\ref{sec:sup_metric}). 
  \textit{Drivable}: drivable-area compliance; \textit{Direct}: driving-direction compliance; \textit{Comf}: comfort; \textit{Prog}: route progress; \textit{Coll}: no at-fault collision factor; \textit{Lim}: speed-limit compliance; \textit{TTC}: time-to-collision within bound.}

  \centering
  \setlength{\tabcolsep}{3.5pt}
  \setlength{\arrayrulewidth}{0.2pt}
\begin{tabular}{c >{\columncolor{lightgray!50}}c ccccccc}
    \toprule
    Method                              & Score & Drivable & Direct. & Comf. & Prog. & Coll. & Lim. & TTC   \\
    \midrule
    UrbanDriver \cite{scheel2022urban}                         & 35.35 & 75.53    & 97.12     & \textbf{98.56}    & \textbf{85.23}   & 55.21      & 81.62       & 47.84 \\
    GCPGP \cite{hallgarten2023prediction}                              & 36.85 & 81.29    & 98.20      & 77.33   & 46.96    & 72.30       & 97.92       & 68.34 \\
    IDM \cite{treiber2000congested}                               & 53.07 & 84.94    & 98.02     & 83.15   & 64.79    & 74.01      & 96.38       & 60.57 \\
    GameFormer \cite{huang2023gameformer}                          & 62.05 & 93.54    & 98.74     & 83.15   & 66.27    & 86.02     & 98.19       & \underline{74.55} \\
    PDM-Hybird \cite{dauner2023parting}                         & 64.05 & 95.34    & 99.10      & 75.98   & 67.93    & \textbf{87.81}       & \textbf{99.57}        & 72.75 \\
    PDM-Closed \cite{dauner2023parting}                         & 64.18 & \underline{95.69}    & \underline{99.10}       & 77.06   & 68.20     & \textbf{87.81}       & \textbf{99.57}        & 73.47 \\
    VLM-planner\cite{tang2025vlmplanner}                          & \underline{66.66} & 90.91    & 95.65       & 84.58   & \underline{84.29}     & 76.88       & \underline{99.37}        & 70.36 \\
    \midrule
    AsyncDriver                        & 65.00  & 94.62 & 98.75 & 83.15 & 67.13 & 85.13 & 98.15 & 73.48 \\ 
    \textit{ours}                  & \textbf{67.28}  & \textbf{96.40}   & \textbf{99.27}   & \underline{86.66}   & 68.16     & \underline{86.99}     & 98.14       & \textbf{75.93}   \\
  \bottomrule
  \end{tabular}
  \label{tab:mainres}
\end{table*}

Compared with VLM-planner, ASSCG has lower route progress but improves the safety-related and rule-compliance terms (drivable area, driving direction, comfort, collision, and TTC), yielding a higher composite score. This suggests that VLM-planner is more aggressive on Hard20, whereas ASSCG trades some progress for more stable closed-loop behavior.



\subsection{Results on NAVSIM (RecogDrive-based Fast--Slow Planner)}
We evaluate ASSCG on NAVSIM using a RecogDrive-based fast--slow instantiation (Appendix~\ref{app:recogdrive_arch}). We keep the original VLM module (InternVL-2B) as the slow branch and construct a lightweight fast branch by replacing the VLM with a ViT backbone; both branches use the same diffusion-planner architecture but are trained as separate policies (no weight sharing). Since NAVSIM is one-shot (predicting a single 4\,s trajectory per scene), we use a simplified ASSCG without buffer/time features and with a binary decision (fast vs.\ slow). To better reflect the sensitivity to future trajectory quality in this setting, we additionally encode the fast-branch predicted trajectory with a lightweight MLP (mapped to a 256-d vector) and feed it to the gate as an auxiliary cue. The NAVSIM gate is trained with RL to directly optimize PDMS.

Table~\ref{tab:navsim_mini} shows a compact comparison. Our RecogDrive-based dual system with ASSCG achieves 91.4 PDMS, outperforming the original ReCogDrive (90.8) and reducing average wall-clock inference latency from $\sim$350\,ms to $\sim$270\,ms on a single A30 GPU. Although the system maintains two trained branches, only one branch is executed per scene after gating in this one-shot setting, so the reported wall-clock latency includes the gating overhead and selected branch inference. Full NAVSIM comparisons are provided in Appendix~\ref{app:navsim}.



\begin{table*}[t]
    \centering
    \small
    \caption{Compact NAVSIM comparison. Full results are in Appendix~\ref{app:navsim}. $^{\ddagger}$Concurrent preprints that appeared after our initial submission.}
    \begin{tabular}{@{}l|cc|ccc|c@{}}
        \toprule
        Method & NC$\uparrow$ & DAC$\uparrow$ & TTC$\uparrow$ & Comf.$\uparrow$ & EP$\uparrow$ & PDMS$\uparrow$ \\
        \midrule
        UniAD~\cite{hu2023planning} & 97.8 & 91.9 & 92.9 & \textbf{100} & 78.8 & \cellcolor{gray!30} 83.4 \\
        TransFuser~\cite{chitta2022transfuser} & 97.7 & 92.8 & 92.8 & \textbf{100} & 79.2 & \cellcolor{gray!30} 84.0 \\
        Hydra-MDP~\cite{li2024hydra} & 98.3 & 96.0 & 94.6 & \textbf{100} & 78.7 & \cellcolor{gray!30} 86.5 \\
        ReCogDrive~\cite{li2025recogdrive} & 97.9 & 97.3 & 94.9 & \textbf{100} & 87.3 & \cellcolor{gray!30} 90.8 \\
        DiffusionDriveV2~\cite{zou2025diffusiondrivev2}$^{\ddagger}$ & \textbf{98.3} & 97.9 & 94.8 & 99.9 & \textbf{88.0} & \cellcolor{gray!30} 91.2 \\
        \midrule
        \textit{ours} & 98.2 & \textbf{98.3} & 94.8 & \textbf{100} & 87.5 & \cellcolor{gray!30} \textbf{91.4}  \\
        \toprule
    \end{tabular}
    \label{tab:navsim_mini}
\end{table*}

\begin{table}[t]
\centering
\caption{Performance and per-frame inference latency.}
\label{table2}
\begin{tabular}{lcc}
\toprule
Method & Score $\uparrow $ & Latency (s/frame) $\downarrow  $ \\
\midrule
AsyncDriver & 65.00 & 0.80 \\
AsyncDriver (5-frame interval) & 64.27 & 0.32 \\
\hline
AdaptiveAsyncDriver (ours) & \textbf{67.28} & \textbf{0.32} \\
\bottomrule
\end{tabular}
\end{table}

\subsection{Ablation Study}
\label{sec:ablation}
Table~\ref{table3} reports ablations on training strategy and backbone choice for AdaptiveAsyncDriver.  

\textbf{Training strategy. }Supervision-only gating achieves 66.21\%. Adding on-policy fine-tuning with GRPO improves the score to 67.28\%, indicating that, without access to optimal labels, post-SFT RL can outperform the fixed-interval surrogate by exploring and refining schedules, suppressing redundant calls in equivalent intervals, and skipping low-yield windows. Replacing GRPO with PPO yields 66.76\%, but we observe instability and mode collapse, likely due to the value network's difficulty in fitting the composite, rule-based score and estimating long-horizon returns.

\textbf{Backbone choice.} Replacing the RWKV backbone in ASSCG with a Transformer block of similar parameter count gives comparable accuracy (67.21\% vs. 67.28\%), but latency scales worse: Transformer attention is quadratic in context length $L$ ($O(L^2)$ time/memory per layer), while RWKV scales linearly in time and $O(1)$ in memory per step after state caching. For $L=150$ (covering a full scene), last-frame gate latency is $\sim$0.06\,s for Transformer vs. 0.02\,s for RWKV on the same A30 GPU. We report \emph{last-frame} gate latency as a practical worst-case proxy: in long-horizon episodes the gate runs with the maximum context length, which upper-bounds the additional per-step overhead and is more relevant for real-time responsiveness than average latency. Although windowing or truncation can reduce $L$, real driving often exhibits variable and long temporal contexts; RWKV’s linear scaling thus provides a more robust latency profile, making it our default backbone.

Additional ablations (e.g., query-cost weight $\lambda_{\mathrm{call}}$, removing \textsc{Drop}, and \textsc{Drop} usage statistics), qualitative visualizations, and our RecogDrive-based NAVSIM instantiation are provided in the Appendix~\ref{app:extra}. In particular, removing \textsc{Drop} lowers the score from 67.28 to 66.71 despite a similar query interval, showing that ASSCG does more than reduce query frequency: it can intentionally suppress harmful slow guidance.


\begin{table}[t]
\centering
\caption{Ablation study of AdaptiveAsyncDriver}
\label{table3}
\begin{tabular}{lcccc}
\toprule
AsyncDriver & SFT& PPO & GRPO & Score $\uparrow $ \\
\midrule
\checkmark & & & &65.00 \\
\checkmark & \checkmark & & &66.21 \\
\checkmark & \checkmark & \checkmark & &66.76 \\
\checkmark & \checkmark & & \checkmark &67.28 \\
\bottomrule
\end{tabular}

\begin{tabular}{lcc}
\toprule
Backbone & Score $\uparrow $ & Last-frame latency(s) $\downarrow $
 \\
\midrule
RWKV block & 67.28 & 0.02 \\
Transformer block & 67.21 & 0.06 \\
\bottomrule
\end{tabular}
\end{table}

\section{Conclusion}
In this work, we analyzed limitations of existing coordination schemes for fast--slow planning and introduced concepts such as the \textit{Equivalent Interval} and \textit{Failure Interval}. Building on this analysis, we proposed the Adaptive Slow-System Control Gate (ASSCG), an architecture-agnostic control interface that schedules and regulates interactions between a learning-based fast planner and an LLM-based slow system. We validated ASSCG on two representative fast--slow instantiations and benchmarks---AsyncDriver on nuPlan Hard20 closed-loop evaluation and a RecogDrive-based fast--slow planner on NAVSIM---showing improved performance--efficiency trade-offs (higher scores with fewer or better-timed slow-module calls). A limitation is that our experiments evaluate when to use a slow branch given existing fast--slow systems, rather than proving that LLM reasoning is necessary for every benchmark or scenario. Harder end-to-end long-tail benchmarks are a natural next step.


\section*{Acknowledgements}
Funded by Xiongan AI Institute, Lenovo Research and Wuxi Research Institute of Applied Technologies, Tsinghua University under Grant 20242001120.
%
%
\bibliographystyle{splncs04}
\bibliography{main}

\clearpage
\hypersetup{pageanchor=false}
\setcounter{page}{1}

\appendix
\renewcommand{\theHsection}{appendix.\arabic{section}}
\renewcommand{\theHsubsection}{appendix.\arabic{section}.\arabic{subsection}}
\section{Per-type fixed-schedule grid search on nuPlan Hard20}
\label{app:gridsearch_stride}
We provide detailed results of the fixed-schedule grid search mentioned in \S\ref{sec:analysis}. For each scenario type (type0--type13), we evaluate a family of fixed query schedules parameterized by the query stride $K \in \{1,3,5,10,20,50,100\}$, where $K{=}1$ queries every frame. We also include \textit{Never-Query}, which disables the slow module entirely. For each type, we report the best average score over its Hard20 episodes and the corresponding schedule (Table~\ref{tab:gridsearch_stride}).

\paragraph{How to interpret the results.}
The wide variation in the best stride across types suggests that the usefulness of slow reasoning is highly scenario-dependent. Importantly, this grid search uses a \emph{type-level} constant stride, i.e., the same schedule is applied to all episodes within a type. This restriction likely underestimates what an adaptive policy could achieve, because even within a fixed scenario type the need for slow guidance can change over time and differ across episodes (e.g., due to interaction density, occlusions, or traffic-light timing). Our goal is therefore to learn a \emph{frame-level} gate that adapts query decisions to the evolving context rather than committing to a single hand-designed schedule.

\begin{table}[H]
\centering
\caption{Per-type best fixed slow-module query schedule on nuPlan Hard20 (grid search). $K$ is the fixed query stride; ``Never'' disables the slow module. Scenario types follow the official nuPlan taxonomy and correspond to type0--type13 in Table~\ref{tab:navsim_results}.}
\label{tab:gridsearch_stride}
\small
\setlength{\tabcolsep}{4pt}
\begin{tabular}{lcc}
\toprule
Scenario type & Best score & Best fixed schedule \\
\midrule
Behind long vehicle & 83.53 & $K{=}1$ \\
Changing lane & 69.57 & $K{=}10$ \\
Following lane with lead & 83.66 & $K{=}5$ \\
High lateral acceleration & 63.02 & Never \\
High magnitude speed & 83.77 & $K{=}3$ \\
Low magnitude speed & 50.00 & $K{=}10$ \\
Near multiple vehicles & 6.27 & $K{=}5$ \\
Starting left turn & 62.55 & $K{=}10$ \\
Starting right turn & 66.12 & $K{=}1$ \\
Starting straight traffic lights traversal & 70.99 & $K{=}50$ \\
Stationary in traffic & 84.90 & $K{=}10$ \\
Stopping with lead & 97.30 & $K{=}10$ \\
Traversing pickup/dropoff & 55.37 & $K{=}10$ \\
Waiting for pedestrian & 55.02 & $K{=}20$ \\
\bottomrule
\end{tabular}
\end{table}

\section{Supplementary: RL Fine-tuning Objective and Details}
\label{sec:supp_rl}

\paragraph{SFT pseudo-label construction.}
The SFT labels are not assumed to be optimal frame-level decisions. They are an initialization derived from closed-loop rollouts: (i) for each training scene, we evaluate a fixed pool of query schedules with stride $K \in \{1,3,5,10,20,50,100\}$ plus \textit{Never-Query}; (ii) each rollout is scored by the official nuPlan closed-loop metric; (iii) the highest-scoring rollout for that scene is selected; and (iv) its per-frame \textsc{Query}/\textsc{Cache}/\textsc{Drop} action sequence is used as the pseudo label. This logarithmic stride pool covers dense, medium, sparse, and no-query regimes while keeping simulator cost manageable. The subsequent RL stage is therefore responsible for refining beyond these coarse fixed schedules.

\paragraph{Logged data and behavior policy.}
We construct a dataset $D$ of trajectories by rolling out the SFT-initialized policy. For each step we log $(s_t,a_t,r_t,\log b_t,\text{done})$, where $b(a_t|s_t)$ is the behavior policy used for data collection and $\log b_t=\log b(a_t|s_t)$ is stored for off-policy correction. The behavior policy is an $\varepsilon$-mixture of a temperature-scaled policy and the uniform distribution:
\begin{equation}
b(a|s) = (1-\varepsilon(s))\,\mathrm{softmax}(\mathrm{logits}(s)/T(s)) + \varepsilon(s)\,U(a).
\end{equation}

\paragraph{Reward shaping.}
Per-step rewards are computed by potential-based shaping plus a query-cost penalty, as described in Eq.~\ref{eq:reward_main}. The terminal step additionally receives the progress settlement term so that the shaped return matches the official score up to a policy-invariant constant.

\paragraph{Advantage estimation.}
We use generalized advantage estimation (GAE):
\begin{equation}
\delta_t = r_t + \gamma V_\phi(s_{t+1}) - V_\phi(s_t),\qquad
\hat{A}_t = \sum_{k\ge 0} (\gamma\lambda)^k \delta_{t+k},
\end{equation}
and the $\lambda$-return target $\hat{R}_t = \hat{A}_t + V_\phi(s_t)$ for value regression.

\paragraph{Off-policy correction and clipped policy objective.}
Let $\pi_\theta$ denote the current policy and $\pi_{\mathrm{ref}}$ the frozen SFT reference. We compute the importance ratio
\begin{equation}
\rho_t(\theta) = \frac{\pi_\theta(a_t|s_t)}{b(a_t|s_t)} = \exp(\log \pi_\theta(a_t|s_t) - \log b_t),
\end{equation}
and apply truncation $\tilde{\rho}_t = \min(\rho_t,\rho_{\max})$ to cap variance. The policy objective uses PPO-style clipping with KL regularization and entropy bonus:

\begin{equation}
\begin{split}
\mathcal{L}_{\pi}(\theta) &=
\mathbb{E}_{(s_t,a_t)\sim D}\!\left[
\min\!\Big(\tilde{\rho}_t\,\hat{A}_t,\;
\mathrm{clip}(\tilde{\rho}_t,1-\epsilon,1+\epsilon)\,\hat{A}_t\Big)
\right] \\
&\quad -\beta\,\mathrm{KL}\!\left(\pi_\theta(\cdot|s_t)\,\|\,\pi_{\mathrm{ref}}(\cdot|s_t)\right)
+\eta\,\mathcal{H}\!\left(\pi_\theta(\cdot|s_t)\right).
\end{split}
\end{equation}

\paragraph{Value regression and overall optimization.}
We minimize mean squared error for the value function:
\begin{equation}
\mathcal{L}_{V}(\phi) = \mathbb{E}_{s_t\sim D}\left[(V_\phi(s_t)-\hat{R}_t)^2\right],
\end{equation}
and jointly optimize policy and value parameters by minimizing
\begin{equation}
\min_{\theta,\phi}\; -\mathcal{L}_{\pi}(\theta) + \lambda_V \mathcal{L}_{V}(\phi).
\end{equation}

\paragraph{Hyperparameters.}
Unless otherwise stated: $\gamma{=}0.99$, $\lambda{=}0.95$, $\lambda_{\mathrm{call}}{=}0.001$, $\epsilon{=}0.2$, $\rho_{\max}{=}10$, $\beta{=}0.01$, $\eta{=}0.01$, and $\lambda_V{=}0.5$. We use AdamW with learning rate $10^{-4}$ and gradient-norm clipping at 1.0.

\section{nuplan Metrics}
\label{sec:sup_metric}
In this section, we elucidate the metrics of closed-loop evaluation deployed within the nuPlan \cite{nuplan} framework. The calculation for the composite metric \textit{Score} is detailed in Formula \ref{equ:nuplan score}.

\noindent \textbf{Drivable Area Compliance (Drivable)} Ego should remain within the designated drivable area. If a frame is identified where the maximum distance from any corner of the ego's bounding box to the nearest drivable area exceeds a predetermined threshold, the drivable area compliance score is assigned a value of $0$; otherwise, it is assigned a value of $1$.

\noindent \textbf{Driving Direction Compliance (Direct.)} The evaluation of the ego's compliance with the correct driving direction is encapsulated within this metric. It stipulates that a distance traversed in the opposite direction of less than $2$ meters within a 1-second interval is deemed compliant, warranting a score of $1$. A traversal exceeding $6$ meters within the same timeframe indicates a significant deviation from compliance, thus yielding a score of $0$. Intermediate distances are assigned a proportional score of $0.5$.

\noindent \textbf{Ego is Comfortable (Comf.)} This metric quantifies the comfort of the trajectory based on a series of kinematic and dynamic thresholds including longitudinal acceleration, lateral acceleration, yaw acceleration, yaw velocity, longitudinal jerk, and overall jerk magnitude. Compliance across all specified parameters results in a trajectory being classified as "comfortable", meriting a score of $1$; otherwise, the score is $0$.

\noindent \textbf{Ego Progress Along Expert Route (Prog.)} This metric represents the ratio of the ego‘s progress per frame to that of an expertly defined route. The closer the ego's progress is to the expert route's progress, the higher the score, with the maximum score being $1$.

\noindent \textbf{No Ego At-fault Collisions (Coll.)} This metric is integral to the safety evaluation, quantifying instances where the ego is directly responsible for collisions. An absence of at-fault collisions throughout the evaluation period scores a $1$. A singular collision with a static object, such as a traffic cone, is deemed a minor infraction, scoring a $0.5$, whereas all other scenarios result in a score of $0$, accentuating the paramount importance of collision avoidance.

\noindent \textbf{Speed Limit Compliance (Lim.)} The ego's adherence to speed regulations is scrutinized by calculating the average instance of speed limit violations per frame. This metric deducts points based on the proximity of the violation to the maximum speed threshold, promoting strict compliance with speed limits.

\noindent \textbf{Time to Collision within Bound (TTC)} This metric evaluates the hypothetical time to collision (TTC) with any external agent if the ego were to continue on its trajectory with unaltered speed and heading. A TTC exceeding $0.95$ seconds is considered safe, thus earning a score of $1$. Lower TTC values signify elevated risk, consequently scoring a $0$, highlighting the significance of maintaining a safe buffer from other road users.

\noindent \textbf{Making Progress (MP)} An ancillary metric to the Ego Progress Along Expert Route, it assigns a score of $1$ if the progress exceeds a threshold of $0.2$, otherwise scoring $0$. Though not directly incorporated into the main text due to its derivative nature, it contributes to the final score.

\noindent \textbf{Score} The final scenario score is calculated by combining the above metrics in a predetermined manner, providing a comprehensive assessment of the autonomous vehicle's performance within the evaluated scenarios:

\begin{equation}
\begin{aligned}
Score = &(5\times Prog + 5\times TTC + 4 \times Lim + 2\times Comf)
\\
& \div 16 \times Coll \times Drivable \times MP \times Direct
\end{aligned}
\label{equ:nuplan score}
\end{equation}

\section{Additional nuPlan Results}
\label{app:nuplan_extra}

\subsection{Per-scenario-type results on Hard20}
\label{app:pertype_results}
Table~\ref{tab:nuplan_pertype} reports the per-scenario-type breakdown (type0--type13) on nuPlan Hard20. We include this table in the appendix for completeness and to facilitate comparison with prior work. The overall trend is consistent with the main text: ASSCG improves most scenario types while keeping the compute budget similar to a fixed-stride baseline, suggesting that the benefit comes from better timing and selective suppression of slow-module usage rather than higher query frequency.


\begin{table*}[ht]
\centering
\caption{Per-scenario-type scores on nuPlan Hard20. Types 0--13 follow the official nuPlan taxonomy and correspond to the 14 scenario types. $^{\dagger}$VLM-planner does not report per-type scores; we denote missing entries as ``--''.}

\begingroup

\renewcommand{\arraystretch}{1.2} 
\setlength{\tabcolsep}{1pt} 
\tiny 
\begin{tabular}{@{}l>{\columncolor{gray!20}}ccccccccccccccc@{}}
\toprule
Methods & score & type0 & type1 & type2 & type3 & type4 & type5 & type6 & type7 & type8 & type9 & type10 & type11 & type12 & type13 \\
\midrule
UrbanDriver\cite{scheel2022urban} & 35.35 & 69.39 & 15.78 & 44.59 & 7.42 & 13.70 & 27.14 & 0.00 & 19.44 & 20.80 & 23.61 & 68.33 & 93.16 & 33.78 & 56.39 \\
GCPGP\cite{hallgarten2023prediction} & 36.85 & 59.50 & 35.91 & 33.60 & 29.33 & 42.84 & 17.24 & 4.86 & 32.33 &  8.22 & 36.23 & 71.32 & 80.46 & 36.19 & 23.61 \\
IDM\cite{treiber2000congested} & 53.07 & 70.69 & 44.84 & \textbf{91.54} & 54.08 & 50.22 & 41.71 & 4.76 & 53.97 & 37.53 & 60.33 & 83.97 & 93.03 & 45.77 & 12.44 \\
GameFormer\cite{huang2023gameformer} & 62.05 & 84.32 & 65.78 & 83.62 & 49.03 & 71.79 & 36.80 & 0.00 & 51.76 & 55.03 & 77.24 & 82.83 & \underline{98.24} & 49.41 & 56.16 \\
PDM-Hybrid\cite{dauner2023parting}  & 64.05 & \textbf{87.68} & 69.61 & \underline{87.20} & 49.95 & 82.80 & 41.32 & 4.23 & 54.98 & 51.72 & \underline{82.95} & 80.37 & \textbf{98.40} & 37.77 & \underline{71.99} \\
PDM-Closed\cite{dauner2023parting} & 64.18 & \underline{87.67} & \underline{70.93} & \underline{87.20} & 49.95 & 82.80 & 41.25 & 4.22 & 53.53 & 51.57 & \textbf{82.96} & 80.37 & \textbf{98.40} & 39.95 & \textbf{72.04} \\
VLM-planner\cite{tang2025vlmplanner}\textsuperscript{$\dagger$} & 66.66 & -- & -- & -- & -- & -- & -- & -- & -- & -- & -- & -- & -- & -- & --  \\
\hline
AsyncDriver\cite{chen2024asynchronous}  & 65.00 & 83.53 & 67.24 & 83.14 & \underline{62.89} & \underline{83.28} & \underline{49.26} & \underline{5.94} & \textbf{62.53} & \underline{65.96} & 64.86 & \underline{84.88} & 96.62 & \underline{49.60} & 54.35 \\

ours  & \textbf{67.28} & 83.40 & \textbf{72.00} & 84.50 & \textbf{63.25} & \textbf{84.11} & \textbf{50.02} & \textbf{6.15} & \underline{62.45} & \textbf{66.02} & 73.00 & \textbf{84.91} & 97.24 & \textbf{60.07} & 55.00 \\
\bottomrule
\label{tab:nuplan_pertype}
\end{tabular}
\endgroup
\end{table*}

\subsection{Broader validation and control policies}
\label{app:controls_runtime}
On the larger Val14 split, ASSCG also improves over the all-query AsyncDriver baseline (77.78 vs. 75.41), suggesting that the learned gate is not specific to the small Hard20 subset. Table~\ref{tab:control_policies} compares ASSCG with simple control policies on Hard20. The first frame is always forced to \textsc{Query}, so the invalid case of selecting \textsc{Cache} before any buffer is available has zero occurrence in all experiments.

\begin{table}[t]
\centering
\caption{Additional control policies on nuPlan Hard20. Random scores are averaged over three seeds.}
\label{tab:control_policies}
\setlength{\tabcolsep}{4pt}
\small
\begin{tabular}{lcc}
\toprule
Policy & Score $\uparrow$ & Avg. query interval \\
\midrule
All \textsc{Query} (AsyncDriver) & 65.00 & 1.00 \\
All \textsc{Drop} (GameFormer) & 62.05 & -- \\
First \textsc{Query}, then always \textsc{Cache} & 63.71 & 149.0 \\
Random uniform actions & 62.08 & -- \\
Random matched action ratio & 62.79 & -- \\
ASSCG & \textbf{67.28} & 5.26 \\
\bottomrule
\end{tabular}
\end{table}

The learned action ratio is approximately \textsc{Cache}:\textsc{Query}:\textsc{Drop}$\approx 6:2:1$, and the average query intervals across the 14 scenario types are 3.3, 3.1, 15.0, 6.6, 29.3, 3.2, 9.2, 4.5, 7.9, 3.3, 26.8, 4.2, 4.4, and 4.7 frames. This spread supports the motivation for frame-level adaptive control rather than a single global fixed stride.

\paragraph{Runtime and delayed-query robustness.}
AsyncDriver is asynchronous: the fast planner continues to produce controls while slow guidance is being refreshed. The remaining runtime bottleneck is therefore the slow branch, not ASSCG itself. In our setting, the theoretical worst case (slow query plus gate overhead) is about 0.82\,s, while the slowest scene averages about 0.44\,s/frame. To probe delayed slow responses, we delay every \textsc{Query} by 3 frames, corresponding to the maximal offset under an approximate 3:1 slow/fast runtime ratio. The score drops mildly from 67.28 to 66.76, still above the all-query AsyncDriver baseline.

\section{Additional Results on NAVSIM}
\label{app:navsim}

\subsection{Benchmark overview and metric}
\label{app:navsim_metric}
We additionally evaluate on NAVSIM, a planning-oriented benchmark with a non-reactive (pseudo closed-loop) simulator: the ego policy does not affect other agents, and each test case requires predicting a single future trajectory for a short horizon. Following NAVSIM, we report the official Predictive Driver Model Score (PDMS), where higher is better:
\begin{equation}
\label{eq:pdms_app}
\mathrm{PDMS} \;=\; \mathrm{NC} \times \mathrm{DAC} \times
\left(\frac{5\cdot \mathrm{EP} + 5\cdot \mathrm{TTC} + 2\cdot \mathrm{C}}{12}\right),
\end{equation}
where $\mathrm{NC}$ denotes no at-fault collisions, $\mathrm{DAC}$ drivable-area compliance, $\mathrm{EP}$ ego progress, $\mathrm{TTC}$ time-to-collision within bound, and $\mathrm{C}$ comfort.

\subsection{RecogDrive-based fast--slow instantiation}
\label{app:navsim_system}
ReCogDrive~\cite{li2025recogdrive} combines a vision-language model (InternVL-2B) with a diffusion-based planner. To construct a fast--slow system, we keep the original InternVL-2B branch as the \emph{slow} module and build a \emph{fast} branch by replacing the VLM with a lightweight vision-only backbone (ViT-large). Both branches use the same downstream diffusion-planner architecture but are trained as separate policies (no weight sharing). We then apply ASSCG to select between fast and slow outputs.

\subsection{ASSCG adaptation for one-shot prediction}
\label{app:navsim_gate}
Unlike nuPlan closed-loop planning, NAVSIM evaluates one-shot 4\,s trajectory prediction (8 waypoints at 0.5\,s). Therefore, we use a simplified variant of ASSCG: we remove the buffer and time-since-last-call features and collapse the action space to a binary gate (select \textsc{Fast} vs.\ \textsc{Slow}). Since the decision is one-shot, we train the gate with RL to directly optimize the PDMS score (without potential-based shaping).

\subsection{Results}
\label{app:navsim_results}

Table~\ref{tab:navsim_results} reports NAVSIM performance. In addition to PDMS, we report wall-clock inference latency per scene (measured on a single NVIDIA A30). Our RecogDrive-based dual system with ASSCG improves PDMS while reducing average inference latency from $\sim$350\,ms to $\sim$270\,ms.


\paragraph{Deployment cost and slow-branch usage.}
The NAVSIM instantiation stores two trained branches, but after the binary gate only one branch is executed for a scene. On a single A30, the standalone fast branch takes about 97\,ms, the slow ReCogDrive branch about 355\,ms, and the gate about 19\,ms. The learned policy selects the slow branch for about 40\% of scenes; the reported $\sim$270\,ms average is measured end-to-end and includes the gate, selected branch inference, and common pre/post-processing overhead. Replacing the InternVL-based slow branch with a separately tuned Qwen3VL-8B slow branch gives 91.37 PDMS with a $\sim$24.7\% speedup over always using that slow branch, further suggesting that the gating principle is not tied to a single slow model.

\subsection{NavHard results}
\label{app:navhard}
We also evaluate the RecogDrive-based instantiation on NavHard, the harder NAVSIM subset suggested by reviewers. Without additional backbone fine-tuning, ASSCG improves over ReCogDrive and the ViT-only fast branch (Table~\ref{tab:navhard_results}). The absolute scores remain low for all compared VLA/VLM-style methods, especially in stage 2, which indicates that NavHard is a challenging benchmark and that stronger slow-branch modeling is complementary to our gating objective.

\begin{table}[t]
\centering
\caption{NavHard results. Stage-1 and stage-2 scores follow the NavHard protocol.}
\label{tab:navhard_results}
\setlength{\tabcolsep}{4pt}
\small
\begin{tabular}{lccc}
\toprule
Method & Overall $\uparrow$ & Stage-1 $\uparrow$ & Stage-2 $\uparrow$ \\
\midrule
ReCogDrive & 26.2 & 68.9 & 37.7 \\
ReCogDrive-ViT (fast only) & 27.1 & 67.2 & 40.1 \\
DiffusionDrive & 27.0 & 66.7 & 40.5 \\
Curious-VLA & 27.6 & -- & -- \\
ReCogDrive-ASSCG & \textbf{28.0} & 68.7 & 40.2 \\
\bottomrule
\end{tabular}
\end{table}

\begin{table*}[t]
    \centering
    \small
    \caption{NAVSIM results on \textit{navtest} using PDMS.$^{\ddagger}$Concurrent preprints that appeared after our initial submission.}
    \begin{tabular}{@{}l|cc|ccc|c@{}}
        \toprule
        Method & NC$\uparrow$ & DAC$\uparrow$ & TTC$\uparrow$ & Comf.$\uparrow$ & EP$\uparrow$ & PDMS$\uparrow$ \\
        \midrule
        DrivingGPT~\cite{chen2024drivinggpt} & 98.9 & 90.7 & 94.9 & 95.6 & 79.7 & \cellcolor{gray!30} 82.4 \\
        UniAD~\cite{hu2023planning} & 97.8 & 91.9 & 92.9 & \textbf{100} & 78.8 & \cellcolor{gray!30} 83.4 \\
        TransFuser~\cite{chitta2022transfuser} & 97.7 & 92.8 & 92.8 & \textbf{100} & 79.2 & \cellcolor{gray!30} 84.0 \\
        PARA-Drive~\cite{weng2024drive} & 97.9 & 92.4 & 93.0 & 99.8 & 79.3 & \cellcolor{gray!30} 84.0 \\
        DRAMA~\cite{yuan2024drama} & 98.0 & 93.1 & 94.8 & 100 & 80.1 & \cellcolor{gray!30} 85.5 \\
        Hydra-MDP~\cite{li2024hydra} & 98.3 & 96.0 & 94.6 & \textbf{100} & 78.7 & \cellcolor{gray!30} 86.5 \\
        ImagiDrive~\cite{li2025imagidrive} & 98.1 & 96.2 & 94.5 & 100 & 80.5 & \cellcolor{gray!30} 86.9 \\
        DiffusionDrive~\cite{liao2024diffusiondrive} & 98.2 & 96.2 & 94.7 & \textbf{100} & 82.2 & \cellcolor{gray!30} 88.1 \\
        WoTE~\cite{li2025end} & 98.5 & 96.8 & 94.9 & 99.9 & 81.9 & \cellcolor{gray!30} 88.3 \\
        AutoVLA~\cite{li2025drivevla} & 98.4 & 95.6 & 98.0 & 99.9 & 81.9 & \cellcolor{gray!30} 89.1 \\
        DriveVLA-W0~\cite{zhou2025autovla} & 98.7 & 99.1 & 95.3 & 99.3 & 83.3 & \cellcolor{gray!30} 90.2 \\
        ReCogDrive~\cite{li2025recogdrive} & 97.9 & 97.3 & 94.9 & \textbf{100} & 87.3 & \cellcolor{gray!30} 90.8 \\
        WAM-diff~\cite{xu2025wam}$^{\ddagger}$ & \textbf{99.1} & 98.3 & \textbf{96.5} & 99.9 & 84.4 & \cellcolor{gray!30} 91.0 \\
        DiffusionDriveV2~\cite{zou2025diffusiondrivev2}$^{\ddagger}$ & \textbf{98.3} & 97.9 & 94.8 & 99.9 & \textbf{88.0} & \cellcolor{gray!30} 91.2 \\
        \midrule
        \textit{ours} & 98.2 & \textbf{98.3} & 94.8 & \textbf{100} & 87.5 & \cellcolor{gray!30} \textbf{91.4}  \\
        \toprule
    \end{tabular}
    \label{tab:navsim_results}
\end{table*}

\section{Additional Ablations and Qualitative Results}
\label{app:extra}

\subsection{Effect of query-cost weight}
\label{app:lambdacall}
We study the impact of the query-cost weight $\lambda_{\mathrm{call}}$ (Eq.~\ref{eq:reward_main}) on the performance--efficiency trade-off. Table~\ref{tab:lambdacall} reports the overall Hard20 score and the average effective query interval under different $\lambda_{\mathrm{call}}$ values.

\begin{table}[t]
\centering
\caption{Ablation on $\lambda_{\mathrm{call}}$ (nuPlan Hard20).}
\label{tab:lambdacall}
\setlength{\tabcolsep}{6pt}
\begin{tabular}{c|cc}
\toprule
$\lambda_{\mathrm{call}}$ & Score $\uparrow$ & Avg. query interval (frames) $\uparrow$ \\
\midrule
0.0005 & 67.25 & 1.91 \\
0.0010 & 67.28 & 5.26 \\
0.0015 & 66.90 & 5.69 \\
0.0020 & 65.61 & 8.28 \\
\bottomrule
\end{tabular}
\end{table}

\subsection{Confidence feature ablation}
\label{app:confidence}
AdaDrive-style activation mechanisms use scene complexity and confidence to modulate LLM assistance. In our AsyncDriver-based setting, adding a confidence cue to ASSCG changes the score only marginally (67.28 to 67.33) while shortening the average query interval from 5.26 to 4.95 frames. We therefore keep the default gate focused on temporal context, buffer similarity, and elapsed time since the last slow query; confidence may be more useful in architectures where candidate-trajectory confidence is less entangled with the fast planner's own aggregation module.

\subsection{RecogDrive-based fast--slow instantiation on NAVSIM}
\label{app:recogdrive_arch}
\begin{figure}[t]
\centering
\includegraphics[width=\linewidth]{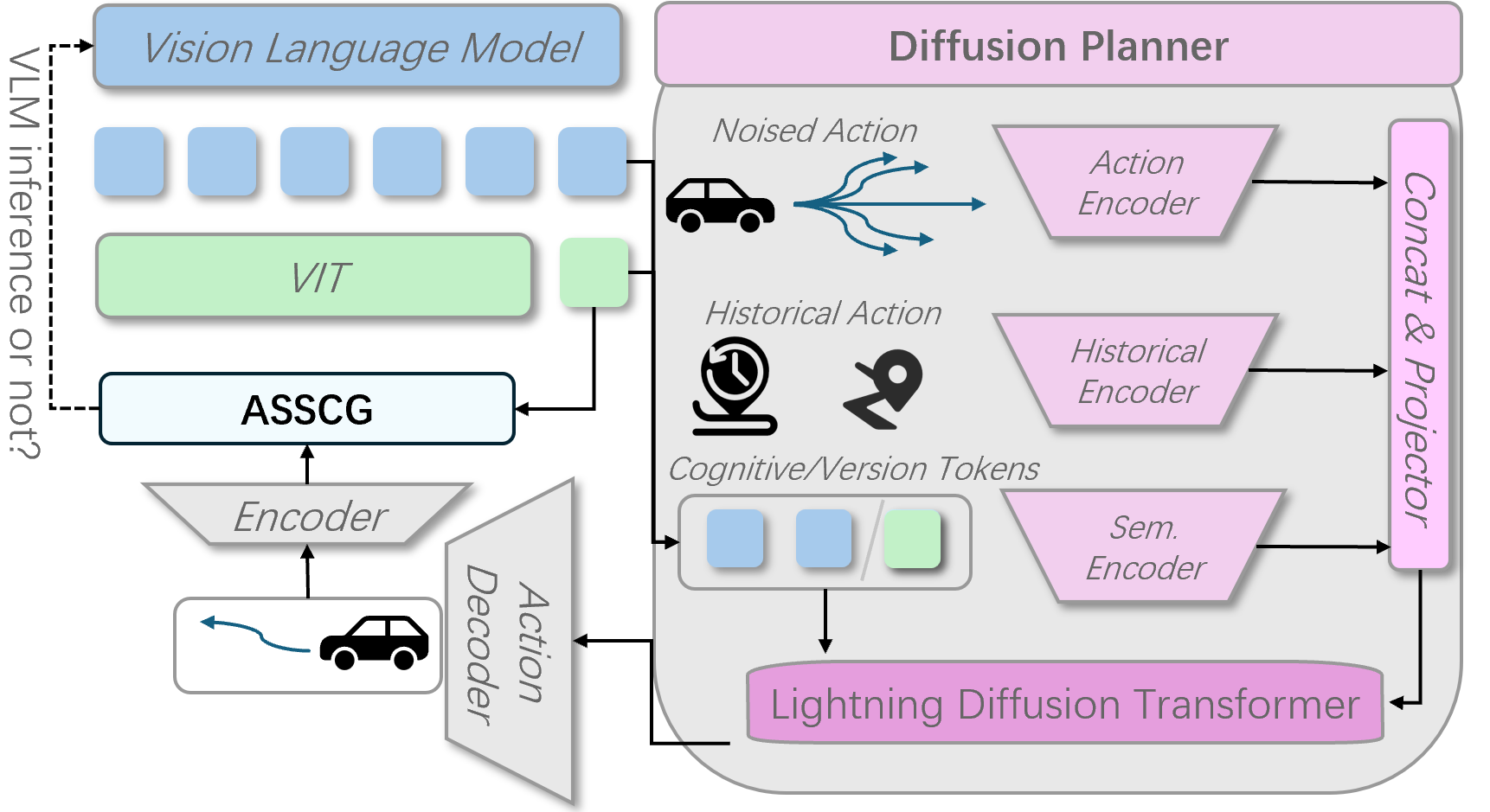}
\caption{RecogDrive-based fast--slow system for NAVSIM. The vision-only fast branch and VLM-based slow branch produce candidate trajectories using the same diffusion-planner architecture; a simplified ASSCG (binary gate) selects the output.}
\label{fig:recogdrive_dual}
\end{figure}

Fig.~\ref{fig:recogdrive_dual} illustrates our RecogDrive-based fast--slow instantiation used on NAVSIM. The slow branch follows the original ReCogDrive design with a VLM (InternVL-2B), while the fast branch replaces the VLM with a lightweight vision-only backbone (ViT-large). Both branches use the same downstream diffusion-planner architecture but are trained as separate policies (no weight sharing). A binary gate selects between the two branches for one-shot trajectory prediction.

\subsection{Additional qualitative visualizations}
\label{app:qual}
\begin{figure*}[t]
\centering
\begin{minipage}[t]{0.48\textwidth}
    \centering
    \includegraphics[width=\linewidth]{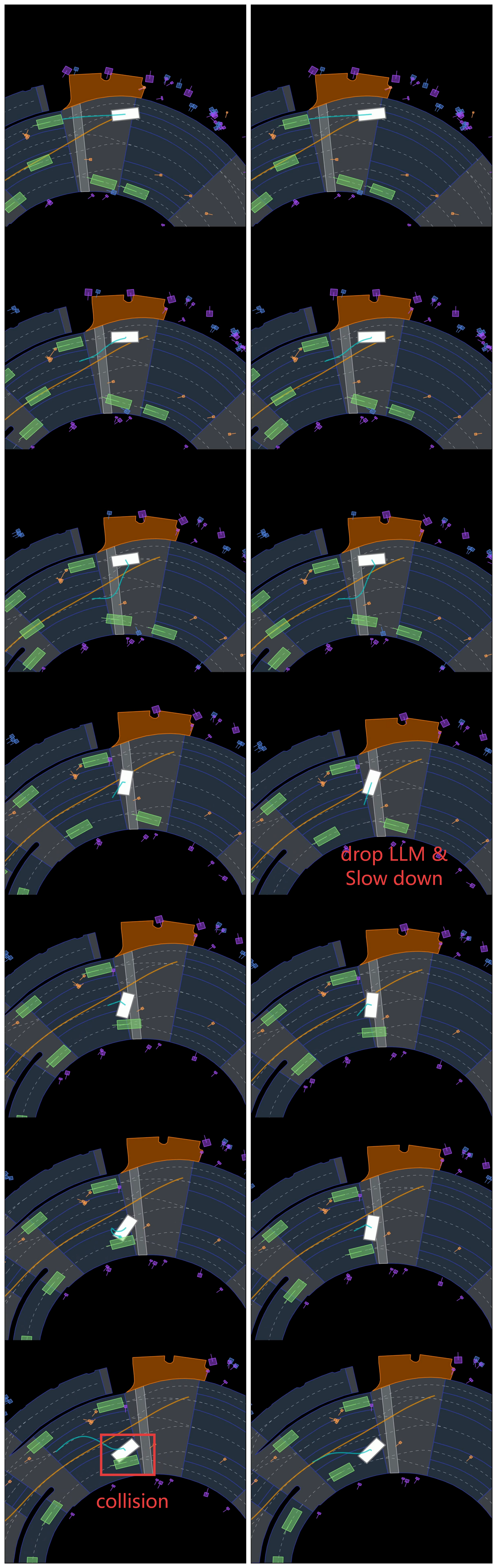}
\end{minipage}
\hfill
\begin{minipage}[t]{0.48\textwidth}
    \centering
    \includegraphics[width=\linewidth]{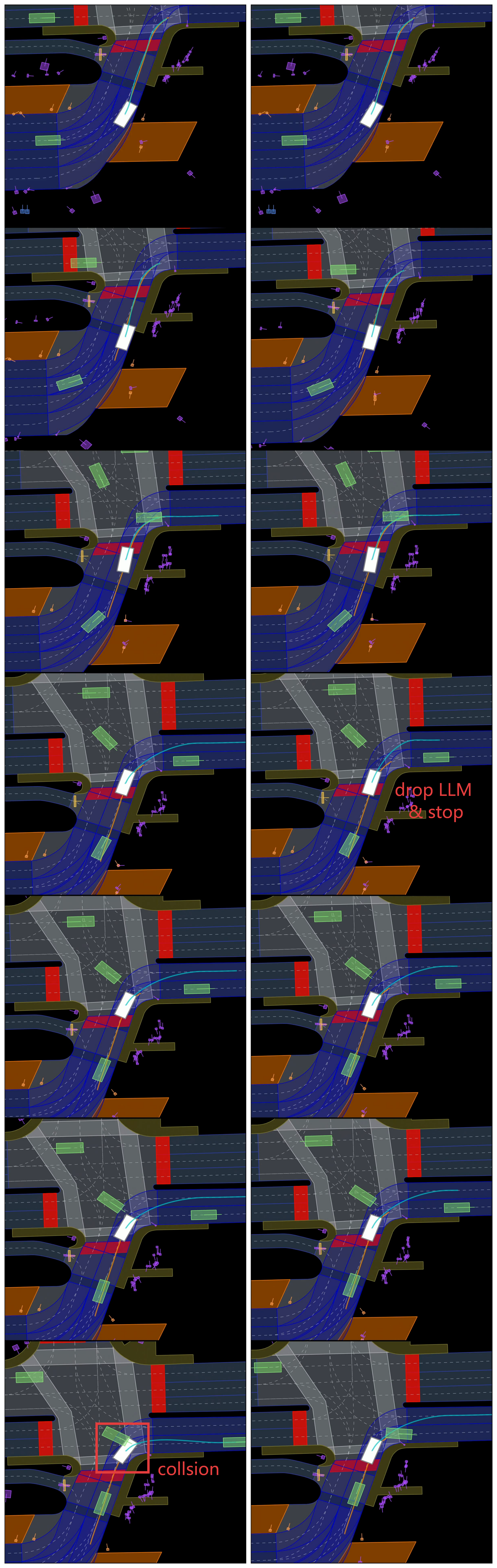}
\end{minipage}
\caption{Additional qualitative examples on nuPlan Hard20.}
\label{fig:qual_cases}
\end{figure*}

We provide two additional qualitative comparisons on nuPlan Hard20 (Fig.~\ref{fig:qual_cases}). Each column corresponds to one scenario (left/right). For each scenario, we visualize 7 synchronized snapshots; the first row shows AsyncDriver (always querying the slow module) and the second row shows our method with ASSCG.

\paragraph{Scenario A (left).}
The 7 frames correspond to steps $\{0,10,30,50,70,90,100\}$. During the first half of the episode (up to $\sim$50), ASSCG issues only three slow-module queries while producing behavior nearly indistinguishable from per-frame querying, illustrating effective reuse within an equivalent interval. Around step $\sim$60, ASSCG outputs \textsc{Drop} to suppress the slow guidance; the ego vehicle decelerates and avoids a subsequent collision.

\paragraph{Scenario B (right).}
The 7 frames correspond to steps $\{0,10,30,50,70,90,110\}$. ASSCG triggers the slow module only 6 times over the entire episode. Around step $\sim$90, ASSCG outputs \textsc{Drop} and the ego commits to a closer-to-stop behavior, preventing a collision with the vehicle on the left.

\subsection{Ablation on the \textsc{Drop} action}
\label{app:drop}

\paragraph{Removing \textsc{Drop}.}
We ablate the \textsc{Drop} action by restricting ASSCG to a 2-way gate (\textsc{Query}/\textsc{Cache}) while keeping all other settings unchanged. As shown in Table~\ref{tab:drop_ablation}, removing \textsc{Drop} reduces the overall score (67.28 $\rightarrow$ 66.71) despite a similar average query interval, indicating that selectively suppressing slow guidance is important beyond merely controlling query frequency.

\begin{table*}[t]
\centering
\caption{Effect of removing \textsc{Drop} on nuPlan Hard20.}
\label{tab:drop_ablation}
\setlength{\tabcolsep}{6pt}
\begin{tabular}{l|cc}
\toprule
Variant & Score $\uparrow$ & Avg. query interval (frames) $\uparrow$ \\
\midrule
ASSCG (Query/Cache/Drop) & 67.28 & 5.26 \\
ASSCG w/o Drop (Query/Cache) & 66.71 & 5.44 \\
\bottomrule
\end{tabular}
\end{table*}

\paragraph{\textsc{Drop} usage statistics.}
Table~\ref{tab:drop_stats} reports how frequently \textsc{Drop} is used. We report (i) the fraction of frames where ASSCG outputs \textsc{Drop} and (ii) the fraction of episodes that contain at least one \textsc{Drop} decision.

\begin{table}[t]
\centering
\caption{\textsc{Drop} usage statistics on nuPlan Hard20.}
\label{tab:drop_stats}
\setlength{\tabcolsep}{6pt}
\begin{tabular}{l|cc}
\toprule
Statistic & Value & Notes \\
\midrule
Drop ratio (frames) & 12 & \% of steps with $a_t=\textsc{Drop}$ \\
Episodes with Drop & 45 & \% of episodes containing at least one \textsc{Drop} \\
\bottomrule
\end{tabular}
\end{table}

\end{document}